\documentclass[10pt,twocolumn,letterpaper]{article}

\usepackage[pagenumbers]{style}

\usepackage{amsmath,amssymb,amsthm}
\usepackage{booktabs,multirow,graphicx,pifont}
\newtheorem{proposition}{Proposition}
\newtheorem{corollary}{Corollary}
\theoremstyle{remark}\newtheorem{remark}{Remark}\theoremstyle{plain}
\newcommand{\efr}{EFR}
\newcommand{\grip}{Strategic training}
\newcommand{\E}{\mathbb{E}}
\newcommand{\Lo}{\mathcal{L}}

\usepackage{cuted}

\usepackage{currfile}

\usepackage{caption}

\definecolor{linkblue}{rgb}{0.21,0.49,0.74}
\usepackage[pagebackref,breaklinks,colorlinks,allcolors=linkblue]{hyperref}
\AtBeginDocument{\crefname{table}{Table}{Tables}\crefname{figure}{Figure}{Figures}}

\title{Beyond Covers and Titles: Evidence-First Ranking for\\Robust Cold-Start Recommendation}

\author{
Fuyi Wang\textsuperscript{1} \quad
Yuxin Cao\textsuperscript{2} \quad
Yifan Liao\textsuperscript{3} \quad
Renyang Liu\textsuperscript{2} \quad
Lin Li\textsuperscript{4} \quad
Wei Song\textsuperscript{5}\thanks{Corresponding author.}
\\[2pt]
\textsuperscript{1}Royal Melbourne Institute of Technology, Australia \quad
\textsuperscript{2}National University of Singapore, Singapore
\\
\textsuperscript{3}The Hong Kong University of Science and Technology (Guangzhou), China
\\
\textsuperscript{4}Southern Cross University, Australia \quad
\textsuperscript{5}Griffith University, Australia
}

\begin{document}
\maketitle
\begin{abstract}
Short-video platforms rank new uploads before anyone has watched them, using content towers that read a video's cover, title and frames. 
Existing attacks on content-aware recommenders perturb pixels or features and ignore platform re-encoding, and existing defenses train against norm-bounded perturbations that neither match real edits nor preserve recommendation quality. We show with REAP, a realizable attack through {a simulated platform channel (H.264 transcoding, rescaling, cropping and JPEG)}, that packaging edits raise a cold video by 32.0\ rank-percentile points in unrelated audiences.
We then observe that cold-start ranking does not need the packaging at all: a tower that sees only the video body ranks cold items best. Building on this, we propose Evidence-First Ranking (\efr), which scores cold videos from their body, admits packaging only through a support-gated residual bounded by $\gamma$, and merges cold and warm rankings by rank percentile, yielding an exact certificate against any packaging edit. On MicroLens-100K, \efr\ cuts the attack gain by 81\% and raises cold-start Recall@100 from 0.1133\ to 0.170, while keeping the merged NDCG@10 within 4\% of the undefended system.
\end{abstract}
\section{Introduction}
\label{sec:intro}

Short-video platforms must rank millions of new uploads before anyone has watched them. For these \emph{cold} videos no interactions exist, so recommenders score them with content towers built on vision-language encoders that read the cover, the title and the frames~\cite{ni2023microlens}. The cover and the title are also the cheapest parts of a video to change: a caption burned into the cover or a few extra hashtags cost nothing and leave the video untouched.

Recommenders that read item images or text are known to be steerable~\cite{liu2021aip,chen2024ipdgi,oh2024atr}, but the risk is usually considered contained. Platforms re-encode every upload, which destroys crafted pixel perturbations~\cite{song2024correction}, and adversarial training against norm-bounded feature perturbations~\cite{tang2018amr,xian2026uatmc} is meant to absorb what survives. We highlight that this view rests on untested premises: attacks are rarely measured after a real codec~\cite{fan2023rethinking}, their gains are not separated into targeted and generic parts, and defenses are tuned to convenient budgets rather than to the size of real edits. We therefore ask three questions:
\textbf{RQ1:} How much exposure can realizable packaging buy after the platform channel?
\textbf{RQ2:} Can robust training remove it without hurting recommendation?
\textbf{RQ3:} How can a platform rank cold videos so that edits to the packaging cannot buy exposure?

{For RQ1, the difficulty is that a creator's edits are neither pixel perturbations nor free feature vectors. They mix discrete choices, such as which caption, sticker or hashtags to add, with continuous ones, such as position, size and color. The creator sees only a black box. Every upload passes a stochastic codec that rescales, crops and recompresses it. We therefore build REAP, which models edits as programs of a grammar of creator operations, with captions proposed by a language model from the cohort's popular titles. It scores each program as an average over sampled draws of the platform channel, so that the search favors edits that survive re-encoding, and it searches in two stages, sampling programs and then refining their continuous parameters with CMA-ES. Because a large gain alone does not show that an edit carries the cohort's meaning, we also run the same search with random words and measure the gain on users outside the target cohort. This decomposition reveals that most of the exposure packaging buys is generic uplift rather than targeted promotion.}

{For RQ2, a fair test of robust training needs a defense that faces the edits creators can actually make. Norm-ball training does not, since real edits move the claim features far beyond the budgets that keep quality. Training on real edits is not enough either: penalizing any score rise after an edit teaches the tower to punish every edited item, including honest ones, and a tower that ignores all changes no longer rewards better content. We therefore design a strategic training that gives robust training its best chance. It trains on the creator's best response, picked by the current model from a cached orbit of realizable edits, asks the target audience not to notice that response through a two-sided term, and keeps real content changes rewarded through a margin. Even this defense, like the published UAT-MC~\cite{xian2026uatmc}, buys robustness only with recommendation quality: a tower that reads the packaging cannot be made insensitive to it without becoming a worse recommender.}

{For RQ3, we observe that cold-start ranking does not need the packaging: a tower that sees only the video body ranks cold items better than one that also reads covers and titles. Turning this into a deployable ranker raises three challenges. The cold score must come from content the creator cannot change for free. Titles and covers still carry some honest signal, so any use of them needs a guarantee that holds for every edit, not only for the edits an attack tried. Cold and warm items are scored by different models whose scores are not on one scale, yet a user sees one list. We propose Evidence-First Ranking (\efr) to meet them. It splits each upload into the creator's claims and the evidence of the video body, with text masked out of the frames, and scores cold items from the evidence alone. The claims can only add a residual of norm at most $\gamma$, gated by their support in the body, so no packaging edit can move any score by more than $\gamma$. Cold and warm items are merged by each model's own rank percentile, which keeps warm rankings intact and carries the $\gamma$ bound over to an exact certificate on the merged list that a user sees.}

Our contributions are summarized as follows:
\begin{itemize} 
\item We introduce REAP, a realizable black-box attack that reveals how creator-editable packaging survives platform processing and systematically inflates the exposure of new videos in cold-start recommendation. 

\item We propose Evidence-First Ranking (EFR), an evidence-driven cold-start ranker that bounds the influence of packaging edits and provides an exact per-user certificate. 

\item We conduct extensive experiments on MicroLens-100K across multiple encoders, ranking architectures, and defenses, showing that EFR substantially reduces packaging attacks while preserving recommendation quality.
\end{itemize}
\section{Background and Threat Model}
\label{sec:background}

\noindent\textbf{Cold-start content towers.}
A sequential recommender scores item $i$ for user $u$ as $s(u,i)=\langle h(u),z_i\rangle$, with $h(u)$ encoding the user's history. A cold item has no interactions, so $z_i$ comes from content: frozen vision-language encoders embed its inputs and a learned tower maps them to $z_i$~\cite{wei2019mmgcn,ni2023microlens}. We split the inputs into the \emph{packaging} or \emph{claims} $C_i$ that the creator presents (cover image, OCR text of the cover, title with hashtags) and the \emph{evidence} $V_i=\{v_1,\dots,v_M\}$, frames sampled over the whole video with text masked, each with a temporal mass $m_k$ equal to its share of the duration.

\noindent\textbf{Attacks on content-aware recommenders.}
Most attacks on recommenders poison interaction data, injecting fake users whose profiles are optimized to promote a target item~\cite{tang2020revisiting,huang2021poisoning,chen2022kgattack}, in the extreme with a single fake user~\cite{huang2023single}. Personalized models trained across clients are likewise exposed to stealthy backdoors. Cold items have no interactions to poison, so the closer threat for them targets content. Perturbed item images promote items in visually-aware recommenders~\cite{dinoia2020taamr,liu2021aip,cohen2021blackbox}, and IPDGI~\cite{chen2024ipdgi} generates such images with guided diffusion. Item text is rewritten against text-aware and LLM-based recommenders~\cite{oh2024atr,zhang2024stealthy,ning2024cheatagent,zhao2026puda}, and VENOMREC~\cite{guan2026venomrec} poisons multimodal LLM recommenders across modalities. The encoders behind these towers are themselves steerable: CLIP reads written words as visual concepts~\cite{materzynska2022disentangling}, and small perturbations move multimodal embeddings to arbitrary targets~\cite{zhang2024illusions}. These attacks perturb pixels, features or text directly and are rarely measured after a video codec. %

\noindent\textbf{Robust multimodal recommendation.}
Adversarial training is the main defense. APR~\cite{he2018apr} and its successors~\cite{zhang2024pamacf} perturb user and item embeddings, AMR~\cite{tang2018amr} perturbs image features, and later work balances modalities, seeks flat minima~\cite{zhong2024mirror} or perturbs hidden layers~\cite{zhao2025dvib}. UAT-MC~\cite{xian2026uatmc} coordinates perturbations across modalities against promotion attacks. Robust CLIP encoders~\cite{schlarmann2024robustclip} harden the encoder instead, adversarial hashing hardens cross-modal retrieval~\cite{ouyang2025ppdacmr}, and adversarial video inputs can be corrected or detected before recognition~\cite{song2024correction}. Other defenses down-weight poisoned interactions~\cite{zhang2024lorec}, detect backdoored updates in federated training~\cite{chen2023fedward,long2024fedcl,chen2023fedcml,shu2026trustworthy}, or certify robustness to bounded feature or data changes~\cite{liu2020certifiable,jia2023pore}. All of them keep reading the packaging, with budgets not tied to realizable edits, and \cref{sec:revisit} shows that their robustness costs quality. Cold-start recommenders map content to collaborative embeddings~\cite{wei2021clcrec,huang2023aldi} but do not separate creator claims from what the video shows. %

\noindent\textbf{Threat model.}
A single creator edits the packaging of their own upload: cover text, stickers, recoloring, crops and hashtags. They cannot write interactions, control other accounts or touch the model. Every edit passes the platform channel $T$: H.264 with a constant rate factor in $[23,35]$, rescaling to 540 to 1080 pixels wide, a 9:16 crop with probability $\tfrac12$, and JPEG with quality in $[60,90]$. %
To favor the attacker, the creator may query the scores of proxy users {(real creators rarely can)} with a budget of $Q$ queries per target. With $\rho(u,i)$ the rank of $i$ for $u$ as a percentage of the catalogue, we define the gain of edit $a$ toward a cohort $g$ that is unrelated to the video as
\begin{equation}
G(a;i,g)=\E_{u\sim g}\,\E_{T}\big[\rho(u,i)-\rho(u,T(a(i)))\big].
\label{eq:gain}
\end{equation}
The platform wants a low gain, high quality for warm and cold items in the merged list, and a continued response to real changes of the video. %
\section{Revisiting Packaging Attacks and Defenses}
\label{sec:revisit}
\label{sec:why}

\subsection{REAP: Realizable Edits through the Platform Channel}
\label{sec:reap}
\noindent\textbf{Edit programs.}
A packaging edit is a program $a=(o,\theta)$ from a grammar of creator operations. Its discrete part $o$ chooses which operations are applied and with what content: a caption burned into the cover, a sticker cut from a popular cover of the cohort, a recoloring toward the palette of such a cover, a crop, and three hashtags appended to the title. Its continuous part $\theta\in[0,1]^{d_o}$ sets the position, size and color of the caption, the position and size of the sticker and the strength of the recoloring. Captions are proposed by Qwen2.5-7B-Instruct from the popular titles of cohort $g$, and hashtags are the TF-IDF keywords of the cohort (\cref{sec:supp_impl}). Applying $a$ to item $i$ changes its claims to $C_i^a$ and leaves the body $V_i$ untouched.

\noindent\textbf{Objective.}
The creator sees neither the platform's users nor the codec draw. With a set $P_g$ of proxy users of the cohort and $K$ independent draws $T_1,\dots,T_K$ of the channel, REAP maximizes the empirical counterpart of \cref{eq:gain},
\begin{equation}
\hat G(a)=\frac{1}{K|P_g|}\sum_{k=1}^{K}\sum_{u\in P_g}\Big[\rho(u,i)-\rho\big(u,T_k(a(i))\big)\Big],
\label{eq:reap}
\end{equation}
under a budget of $Q$ evaluations of $\hat G$. Averaging over channel draws makes the search prefer edits that survive re-encoding, rescaling and cropping over edits that exploit one particular output of the codec~\cite{athalye2018eot}.

\noindent\textbf{Search.}
Because the grammar mixes discrete and continuous choices, REAP searches in two stages. It first samples $N$ programs from the grammar, renders each through the $K$ channel draws and keeps the best, $a_0=(o_0,\theta_0)=\arg\max_{n\le N}\hat G(a_n)$. It then fixes $o_0$ and refines $\theta$ with CMA-ES~\cite{hansen2016cma} for the remaining $Q-N$ queries, $\theta^\star=\arg\max_{\theta}\hat G\big((o_0,\theta)\big)$. We use $N=120$, $K=4$, $|P_g|=500$ and $Q=200$, and score the final edit on 1,000 cohort users disjoint from $P_g$ under fresh channel draws. CMA-ES uses population 8 and initial step 0.2 in normalized coordinates over the continuous parameters.

\noindent\textbf{Decomposing the gain.}
A large gain does not show that an edit carries the cohort's semantics. We therefore run a \emph{random control}, the same search with every caption and hashtag replaced by random words from the title vocabulary, and report the gain over it, together with the gain among users outside the target cohorts. The control keeps the placements, styles and budget of REAP.

\begin{table}[t]
\centering\footnotesize\setlength{\tabcolsep}{1.6pt}
\caption{Attack performance of REAP on undefended towers.}
\label{tab:threat}
\resizebox{0.7\linewidth}{!}{\begin{tabular}{@{}lccc@{}}
\toprule
 & CLIP & SigLIP & Hybrid\\
\midrule
Random control & 26.3  & 25.7  & 12.2  \\
REAP & 34.3   & 32.0 & 19.3  \\
Over control & 8.0   & 6.3   & 7.1  \\
Other-user gain & 26.2 & 25.6 & 9.3\\
Reach (\%) & 1.1$\to$17.0 & 1.5$\to$15.2 & 1.7$\to$7.9\\
\bottomrule
\end{tabular}}
\end{table}

\begin{figure*}[t]
\centering
\includegraphics[width=\linewidth]{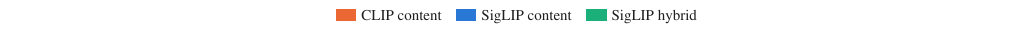}\\[2pt]
\begin{subfigure}[b]{.66\linewidth}\includegraphics[width=\linewidth]{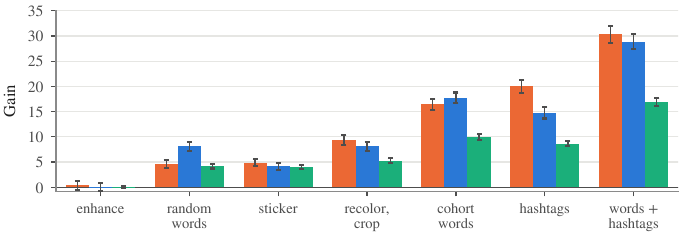}\caption{Realizable edits after the channel}\end{subfigure}\hfill
\begin{subfigure}[b]{.32\linewidth}\includegraphics[width=\linewidth]{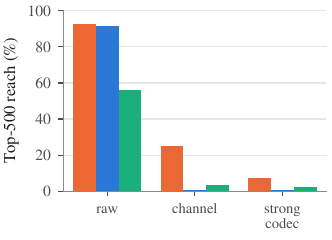}\caption{Pixel PGD, 8/255}\end{subfigure}
\caption{Pilot study of edit families and pixel attacks. %
}
\label{fig:pilot_attack}
\end{figure*}

\begin{table*}[t]
\centering\footnotesize\setlength{\tabcolsep}{3pt}
\caption{Performance of defenses that keep reading the packaging on SigLIP.}
\vspace{-4mm}
\label{tab:main}
\begin{tabular}{@{}lcccccccc@{}}
\toprule
Method & NDCG@10 & Cold R@100 & REAP gain & Over control & Other-user & Reach (\%) & Body change & Retention\\
\midrule
Free (undefended) & 0.0475 & 0.1133 & 29.2 & 4.1 & 24.6 & 1.6$\to$12.8 & 9.8 & 1.00\\
Body & 0.0210 & 0.1697 & 0.0 & 0.0 & 0.0 & 2.9$\to$2.9 & 23.6 & 2.40\\
Cover text masking\textsuperscript{*} & 0.0410 & 0.0989 & 29.6 & n/a & 23.2 & 1.6$\to$12.8 & n/a & n/a\\
Random-edit invariance & 0.0463 & 0.1152 & 23.3 & 1.2 & 21.1 & 1.4$\to$6.1 & 9.8 & 1.00\\
AMR, $\epsilon{=}0.1$ & 0.0304 & 0.1485 & 2.4 & 0.3 & 1.9 & 1.5$\to$1.7 & 24.6 & 2.51\\
UAT-MC, $\epsilon{=}0.1$ & 0.0413 & 0.1418 & 16.5 & 3.5 & 14.2 & 1.1$\to$3.4 & 19.3 & 1.96\\
\grip & 0.0409 & 0.1135 & 16.4 & 0.1 & 15.6 & 1.5$\to$3.7 & 14.2 & 1.44\\
Strategic + bottleneck & 0.0376 & 0.0991 & 13.3 & -0.8 & 12.8 & 1.4$\to$2.3 & 11.9 & 1.21\\
\bottomrule
\end{tabular}
\end{table*}

\subsection{Observations}
\label{sec:obs}

\noindent\textbf{Observation 1: packaging survives the channel and acts mostly as generic uplift.}
We first measure how much exposure realizable packaging buys after the platform channel (RQ1). REAP raises a cold video by 34.3, 32.0\ and 19.3\ points on CLIP, SigLIP and hybrid towers (\cref{tab:threat}), and its reach, the share of cohort users with the item in their top 500, grows from 1.5\% to 15.2\% on SigLIP. A white-box pixel attack falls from 91.5\% to 0.6\% reach after the channel on SigLIP (\cref{fig:pilot_attack}b, 60 targets with 2 cohorts each). The pilot of \cref{fig:pilot_attack}a renders about 40 edits per target and cohort through a fixed channel draw (CRF 28, 540 pixels, JPEG 70), chooses the best of each family on 500 proxy users and scores it on 1,000 disjoint users. Cover text and title hashtags are the strongest single edits (15 to 20 points on content towers) and combine to about 30, while benign enhancement gains nothing. The random control obtains 25.7\ of the 32.0\ points on SigLIP, and users outside the target cohorts gain 25.6: the gain comes from searching for any visible edit the tower over-rewards, not from cohort semantics. {Without querying the deployed model, an edit selected on a surrogate tower still gains 24.0 points.}

\noindent\textbf{Observation 2: robust training trades quality for safety.}
We then ask whether robust training can remove this exposure without hurting recommendation (RQ2). We trained the Free tower (\cref{sec:tower}) against the attack in two ways. The first is UAT-MC~\cite{xian2026uatmc}, which perturbs the claim features within an $\ell_2$ ball. Budgets beyond $\epsilon=0.1$ collapse quality, so the validation rule selects $\epsilon=0.1$.

The second is a stronger, attack-aware \emph{strategic training} that we built to give robust training its best chance: it trains on realizable best responses instead of norm balls. Offline, we render an orbit (\cref{sec:orbit}) $\mathcal{O}(i,g)$ of 12 REAP programs for each of 8,000 pairs of a warm item $i$ and an unrelated cohort $g$. Every $R=3$ epochs, the current model selects the best response $a^\star=\arg\max_{a\in\mathcal{O}(i,g)}\sum_{u\in U_g}s(u,a(i))$ over training users $U_g$ of the cohort, and the tower minimizes the following loss
\begin{equation}
\begin{aligned}
\Lo={}&\Lo_{rec}+\lambda_1\,\E_{(i,g)}\Big|\tfrac{1}{|U_g|}\textstyle\sum_{u\in U_g}\big(s(u,a^\star(i))-s(u,i)\big)\Big|\\
&+\lambda_2\,\E\Big[\big(\mu-\langle\bar h_g,\,z(C_i,V_i^{+})-z(C_i,V_i)\rangle\big)_+\Big].
\end{aligned}
\label{eq:strategic}
\end{equation}
The second term asks the cohort not to notice the best response. It is two-sided (\cref{sec:inv}), because penalizing only a rise teaches the tower to punish every edited item%
. The third term (\cref{sec:imp}) keeps real content rewarded: $V_i^{+}$ replaces half of the timeline with footage of a popular cohort video, and the item must move toward the mean cohort user $\bar h_g$ by a margin $\mu=0.1$. The validation rule selects $\lambda_1=2$ and $\lambda_2=0.5$ for this tower.

Both defenses cut the gain from 29.2\ to about 16 points (\cref{tab:main}), at NDCG@10 changes of -13.8\% for strategic training and -13.1\% for UAT-MC, yet the 95\% intervals of both gains stay above 11 points. AMR at the same budget removes the gain but loses 36\% of NDCG@10. Restricting the strategic tower further, so that claims can only re-weight body frames through a capped attention bottleneck, lowers the gain to 13.3 but costs 21\% of NDCG@10, and its worst-case bound is vacuous (\cref{sec:eb}). {Simpler defenses do not help either (\cref{sec:supp_results}). Masking text on the cover, attacked on the first 100 targets only (\textsuperscript{*} in \cref{tab:main}), leaves the gain at 29.6\ (33.2\ for Cover on the same targets), because titles and hashtags still carry it, and invariance to random edits only lowers it to 23.3, since random edits do not cover the attacker's best response.} As long as the cold item is scored from its packaging, robustness is bought with recommendation quality. The last two columns of \cref{tab:main} check that a defense still responds to real content. Body change is the gain toward the matching cohort when half the timeline is replaced by cohort footage, retention divides it by that of Free, and n/a marks towers without body input.

\noindent\textbf{{Unseen edits and honest creators.}}
This experiment checks whether strategic training merely memorizes REAP's grammar and whether defenses harm honest creators (\cref{tab:probe}). With held-out keywords and fonts, strategic training lowers the gain from 18.5 to 10.0 while AMR and UAT-MC remain at the undefended level. However, every tower that reads the packaging also penalizes honest creators: when a creator simply repeats their own title on the cover, the item's rank drops by 6.7 points under strategic training and by 25--40 points under a one-sided penalty, which effectively learns to punish any edited item. {This tests a single honest edit. Honest subtitles, crops and color grading remain untested.}

\noindent\textbf{{Ablating strategic training.}}\label{sec:exp_ablation}
{To verify that this baseline is not weakened by a design choice, \cref{tab:ablation} removes one part at a time from seed 0 (seeds span 13.3 to 19.6, so only gaps above six points count). Without the strategic terms the gain returns to the undefended level, so the bottleneck alone is no defense. A one-sided penalty yields negative gains even outside the cohorts, as the tower marks every edited item down. A listwise KL term leaves 9.4 more points than the two-sided term on the bottleneck variant.}

\noindent\textbf{Observation 3: cold-start ranking does not need the packaging.}
{Finally, we ask what the packaging contributes to cold-start ranking. In \cref{tab:main}, a tower that reads only the body reaches a cold-start Recall@100 of 0.1697, against 0.1133\ for the same tower with packaging and at most 0.1485\ (AMR) for any defense that keeps reading it.} It is a poor warm ranker (NDCG@10 0.0210\ against 0.0475), also on warm items with whole-video frames (0.014 against 0.040). The packaging helps warm items and hurts cold ones, which is the asymmetry \efr\ exploits.

\begin{table}[t]
\centering\footnotesize\setlength{\tabcolsep}{3pt}
\caption{Unseen edits and honest creators on SigLIP.}
\vspace{-4mm}
\label{tab:probe}
\resizebox{\linewidth}{!}{\begin{tabular}{@{}lccc@{}}
\toprule
Model & Held-out gain & Reach (\%) & Honest edit\\
\midrule
Cover & 18.5 & 1.1$\to$4.7 & $-8.5$\\
Free & 17.9 & 1.0$\to$5.6 & $-10.3$\\
Random-edit invariance & 18.8 & 1.0$\to$4.4 & $-1.8$\\
AMR & 18.9 & 1.0$\to$3.1 & $-0.9$\\
UAT-MC & 19.5 & 1.0$\to$3.1 & $-0.6$\\
\midrule
\grip & 10.0 & 1.1$\to$2.0 & $-6.7$\\
Strategic + bottleneck & 8.2 & 0.7$\to$0.9 & $-6.1$\\
\grip, one-sided & $-22.7$ & 1.2$\to$0.1 & $-39.6$\\
\bottomrule
\end{tabular}}
\end{table}
\section{Our Defense: Evidence-First Ranking}
\label{sec:efr}

\subsection{Design Challenges and Overview}
\label{sec:efr_overview}

\Cref{sec:revisit} suggests removing the packaging from the cold path rather than training a tower to resist it. Turning this into a ranker that a platform can deploy raises three challenges.

\begin{itemize}
    \item \textbf{Challenge I: what should a cold item be scored from?} The score must come from content the creator cannot change for free, and it must still rank cold items well.
    
    \item \textbf{Challenge II: how much may the packaging count, with a guarantee?} Titles and covers carry some honest signal, such as a language or a topic the frames do not show, so ignoring them entirely wastes information. Any use of them, however, must come with a bound that holds for every edit, not only for the edits an attack tried.
    
    \item \textbf{Challenge III: how do cold and warm items share one list?} A user sees one list, while cold and warm items are scored by different models whose scores are not on one scale. The merge must keep warm quality and must not void the guarantee of Challenge II.
\end{itemize}

\begin{figure*}[t]
\centering
\includegraphics[width=\linewidth]{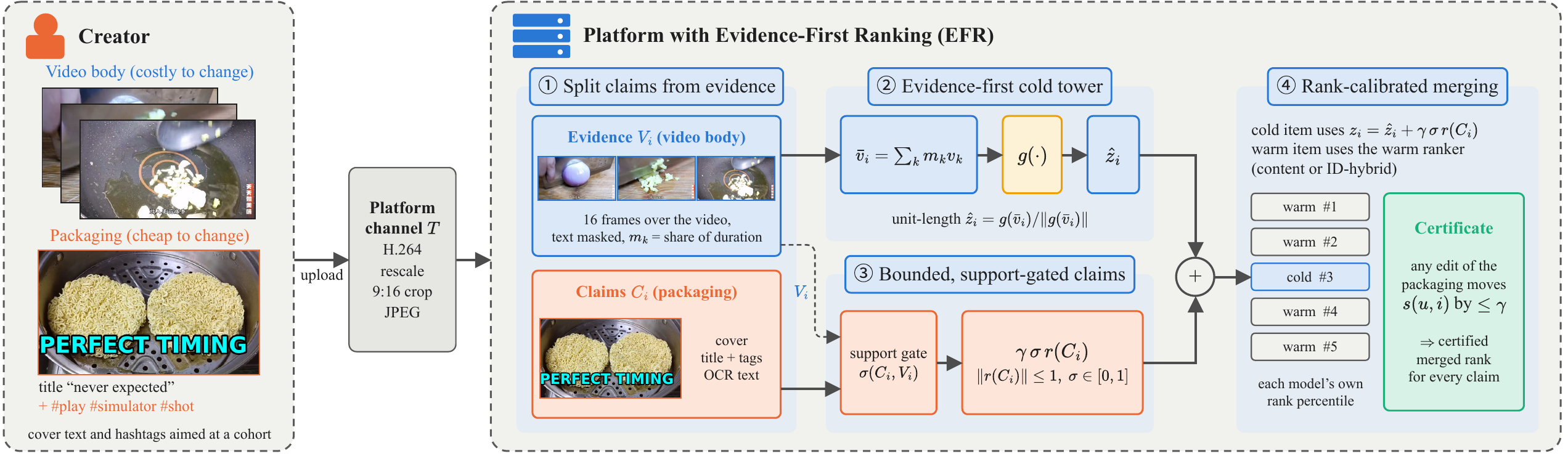}
\caption{Overview of Evidence-First Ranking (\efr). \efr\ scores cold videos by the body, caps packaging at $\gamma$ and merges by rank.}
\label{fig:overview}
\end{figure*}

\noindent\textbf{Overview.}
\Cref{fig:overview} illustrates how \efr\ addresses the three design challenges while preserving the platform's existing warm-item ranker. A creator first uploads a video together with its packaging. In the example shown, a real REAP edit of a cold cooking video targeting the Game cohort adds a caption and three hashtags, increasing its rank by 43 percentile points on the undefended tower. After the platform channel re-encodes the upload, \ding{172} \efr\ splits the creator's claims from the evidence in the video body (Challenge I, Sec.~\ref{sec:efr_split}). \ding{173} A cold tower embeds the evidence into a unit vector and computes the cold score solely from the video body (Challenge I, Sec.~\ref{sec:efr_tower}). \ding{174} The claims can influence the score only through a support-gated residual with norm at most $\gamma$, which bounds the effect of any packaging edit (Challenge II, Sec.~\ref{sec:efr_res}). \ding{175} Cold and warm items are merged using each model's own rank percentile, preserving warm-item rankings and carrying the $\gamma$ bound to the final recommendation list, yielding an exact certificate on the effect of arbitrary packaging edits (Challenge III, Sec.~\ref{sec:efr_merge}).

\begin{table}[t]
\centering\footnotesize\setlength{\tabcolsep}{1.9pt}
\caption{Ablations of strategic training on SigLIP.}
\vspace{-4mm}
\label{tab:ablation}
\resizebox{0.9\linewidth}{!}{\begin{tabular}{@{}lcccccc@{}}
\toprule
 & \multicolumn{3}{c}{\grip} & \multicolumn{3}{c}{Strategic + bottleneck}\\
\cmidrule(lr){2-4}\cmidrule(l){5-7}
Variant & NDCG & Gain & Title & NDCG & Gain & Title\\
\midrule
Full method & 0.0430 & 19.6 & 3.6 & 0.0391 & 14.8 & 1.1\\
No $\Lo_{inv}$, no $\Lo_{imp}$ & 0.0477 & 29.3 & 1.3 & 0.0437 & 29.8 & 6.3\\
No $\Lo_{imp}$ & 0.0431 & 18.9 & 1.2 & 0.0402 & 16.8 & -0.8\\
Random orbit candidate & 0.0435 & 23.6 & 0.7 & 0.0378 & 18.9 & 1.5\\
Listwise KL invariance & 0.0486 & 23.2 & -0.6 & 0.0430 & 24.2 & 6.4\\
One-sided invariance & 0.0478 & -15.4 & -0.4 & 0.0430 & -7.8 & 0.3\\
No temporal mass & n/a & n/a & n/a & 0.0394 & 16.5 & 0.5\\
\bottomrule
\end{tabular}}
\end{table}

\subsection{Splitting Claims from Evidence}
\label{sec:efr_split}
After the platform channel, \efr\ separates what the creator \emph{claims} about the video from what the video \emph{shows}. The claims $C_i=(c_{\mathrm{img}},c_{\mathrm{txt}},c_{\mathrm{ocr}})$ are the embeddings of the cover image, of the title with its hashtags and of the text read from the cover by OCR. The evidence $V_i=\{v_1,\dots,v_M\}$ consists of $M=16$ frames sampled uniformly over the whole video. Before encoding, a text detector finds text regions in every frame and fills them with a blurred background, so a caption burned into the frames cannot re-enter as evidence. Each frame carries a temporal mass $m_k$, its share of the duration, so a short insert counts only for the time it occupies. Claims and evidence are embedded with the same frozen vision-language encoder.

\subsection{Evidence-First Cold Tower}
\label{sec:efr_tower}
Observation 3 shows that the body alone ranks cold items better than the packaging does. The cold tower therefore embeds the mass-weighted evidence mean $\bar v_i=\sum_k m_k v_k$ with a two-layer MLP $g$ and normalizes the result:
\begin{equation}
\hat z_i=\frac{g(\bar v_i)}{\|g(\bar v_i)\|}.
\label{eq:base}
\end{equation}
It is trained with the same user tower and full-softmax next-item loss as any content tower, on warm items only, so it never sees a cold item during training. The unit norm keeps the cold score from encoding popularity, a signal that cold items cannot have by definition.

\subsection{Bounded, Support-Gated Claims}
\label{sec:efr_res}
\efr\ admits the packaging through a bounded residual:
\begin{equation}
z_i=\hat z_i+\gamma\,\sigma(C_i,V_i)\,r(C_i),\quad \|r(C_i)\|\leq1,\quad \sigma\in[0,1].
\label{eq:efr}
\end{equation}
Here $r$ is an MLP of the claim embeddings whose output is rescaled to norm at most one, and $\sigma$ is a support gate: a small network on the cosines between each claim and the evidence frames in the shared encoder space, which silences claims that match nothing in the body. The gate decides how the budget $\gamma$ is spent. The guarantee below does not depend on it. {The certificate comes from the norm bound alone. The gate only shapes utility and the empirical gain.}

\begin{proposition}[Packaging bound]
\label{prop:efr}
Let $h(u)$ be a unit user vector and fix the evidence $V_i$. For every packaging $C$ of item $i$ and every edit $a$ of this packaging,
\begin{equation}
\begin{aligned}
&\big|s(u,i;C)-\langle h(u),\hat z_i\rangle\big|\leq\gamma,\quad\text{hence}\\
&s(u,a(i))\leq\langle h(u),\hat z_i\rangle+\gamma=:\bar s(u,i).
\end{aligned}
\end{equation}
\end{proposition}
\begin{proof}
$s(u,i;C)-\langle h(u),\hat z_i\rangle=\gamma\,\sigma\,\langle h(u),r(C)\rangle$, and $|\langle h(u),r\rangle|\leq\|h(u)\|\,\|r\|\leq1$ with $\sigma\in[0,1]$.
\end{proof}

\Cref{prop:efr} holds for every string, sticker and color, including ones no attacker would try. It is attained for a user whenever the gate is fully open and the residual points along $h(u)$, so it relaxes nothing about the architecture, unlike bounds on attention bottlenecks. Its only slack is that one upload cannot point its residual at every user at once. The bound also leaves the attacker no better operation. Against \efr\ at $\gamma=0.02$, REAP still puts captions in 97\% of its programs, against 94\% on Free (\cref{fig:edit_mix}).

\subsection{Rank-Calibrated Merging and Certificate}
\label{sec:efr_merge}
Scores of the two models are not on the same scale, so \efr{} merges rankings rather than raw scores. The warm-item ranker and the cold tower each rank the entire catalogue independently for a given user. A warm item enters the final list according to its rank percentile under the warm ranker, whereas a cold item enters according to its rank percentile under the cold tower. The final list merges these percentiles. Both percentiles are uniform over the catalogue, so neither model fills the top of the list only because its scores run larger. This procedure preserves the relative ordering of warm items. Moreover, because the merging rule is monotonic with respect to the cold-item score, the guarantee of Proposition~\ref{prop:efr} carries over to the final ranking.

\begin{corollary}[Merged-rank certificate]
\label{cor:efr}
For every packaging edit of a cold item $i$, its merged rank for user $u$ is no better than the merged rank of a cold item with cold-tower score $\bar s(u,i)$. Averaging over a cohort gives a certified upper bound on the gain of \cref{eq:gain} over all packaging edits.
\end{corollary}

\noindent\textbf{Choosing $\gamma$.}
$\gamma$ is an explicit knob between how much the packaging can help honest creators and how much it can help dishonest ones, and \cref{prop:efr} turns it into a guarantee a platform can state in advance. $\gamma=0$ recovers a router that scores cold items from the body alone. We fixed $\gamma=0.1$ before seeing any result of \efr\ and report the whole range $\gamma\in\{0.02,0.05,0.1,0.2\}$, since each platform will choose its own operating point.
\section{Extensive Evaluation}
\begin{table*}[t]
\centering\footnotesize\setlength{\tabcolsep}{4pt}
\caption{Defense performance of \efr\ in the merged list on SigLIP. \textsuperscript{\dag}Attacked on fewer seeds than were trained.}
\vspace{-4mm}
\label{tab:efr}
\begin{tabular}{@{}lcccccc@{}}

\toprule \multirow{2}{*}{\textbf{System}} & \multicolumn{2}{c}{\textbf{Utility}} & \multicolumn{4}{c}{\textbf{Robustness}} \\ \cmidrule(lr){2-3} \cmidrule(lr){4-7} & NDCG@10 & Cold R@100 & REAP gain & Reach (\%) & Certified gain & Certified $<5$ (\%)\\ \midrule
Free (undefended) & 0.0469 & 0.1133 & 29.2 & 1.6$\to$12.8 & -- & --\\
\midrule
Router, body only ($\gamma{=}0$) & 0.0454  & 0.1637 & 0.0 (by construction) & unchanged & 0.0 & 100\\
\efr, $\gamma{=}0.02$ & 0.0451 & 0.1699 & 5.4 & 3.3$\to$5.0 & 7.5 & 31\\
\efr, $\gamma{=}0.05$ & 0.0452 & 0.1684 & 10.9 & 2.9$\to$6.1 & 16.4 & 8\\
\efr, $\gamma{=}0.1$  & 0.0458 & 0.1556 & 17.4 & 3.0$\to$9.2 & 27.7 & 1\\
\efr, $\gamma{=}0.2$ & 0.0465 & 0.1408 & 23.4\textsuperscript{\dag} & 2.4$\to$9.6 & 38.8 & 0\\

\bottomrule
\end{tabular}
\end{table*}

\subsection{Experimental Setup}
\label{sec:protocol}
\noindent\textbf{Data.} {We use two datasets from different short-video platforms.} MicroLens-100K~\cite{ni2023microlens} (100,000 users, 19,738 videos, 719,405 comments). We hold out 1,000 cold items and remove their interactions from training. Warm items use leave-last-out splits, giving 38,776 cold-start cases. All cold items and about 2,000 warm items have sixteen frames over the whole video. Other warm items have five frames from the first seconds. {The second, which we call ShortVideo, is the dataset of Shang et al.~\cite{shang2025shortvideo} (3,687 users, 8,779 videos, 34,058 positive interactions, first frame as the cover, 500 cold items, \cref{sec:supp_impl}). Results are on MicroLens-100K unless we state otherwise.}

\noindent\textbf{Targets.} 300 cold items from small tags, each attacked toward the three largest interest cohorts (Game, Anime, Film and TV), i.e., 900 target-cohort pairs. Two further cohorts (Delicacy, Daily Sharing) are held out from training to test generalization to unseen cohorts.

\noindent\textbf{Rankers.} A SASRec-style user tower~\cite{kang2018sasrec} over frozen CLIP ViT-B/32 or SigLIP so400m features, trained with full-softmax next-item loss. \emph{Free} reads packaging and body mean and is the undefended reference. \emph{Body} reads the body only. Hybrid rankers add item IDs for warm items.

\noindent\textbf{Merged list.} \efr\ and the router use Free (or hybrid Free) for warm items and their cold tower for cold items, merged by rank percentile. Single-model defenses rank all items with one tower. We report NDCG@10 on warm test items with cold items competing, and cold-start Recall@100 over the whole catalogue of videos.

\noindent\textbf{Metrics.} We evaluate recommendation quality using NDCG@10 ($\uparrow$), which measures ranking quality on the final merged list, and Cold Recall@100 ($\uparrow$), which measures the fraction of relevant cold-start items retrieved in the top 100 recommendations. To evaluate attack effectiveness, we report REAP gain ($\uparrow$), the mean rise of an edited target in rank percentile among users of an unrelated cohort (\cref{eq:gain}), and Reach ($\uparrow$), the share of cohort users with the target in their top 500 before and after the attack. To evaluate certified robustness, we report Certified gain ($\downarrow$), the largest provable attack gain under the certification budget, and Certified $<5$ ($\uparrow$), the share of target-cohort pairs whose certified gain is below 5 points. Arrows mark the preferred direction.

\noindent\textbf{Implementation.} The user tower is a two-layer causal Transformer with two heads over the last 20 items. Item and user vectors have 128 dimensions. All towers are trained with full-softmax cross-entropy at temperature 0.07, AdamW (learning rate $10^{-3}$, weight decay 0.01), batch size 256 and 20 epochs, keeping the epoch with the best validation NDCG@10. In \efr, $g$ and $r$ are two-layer MLPs of widths 512 and 256, and the gate maps the maximum and the mass-weighted cosine of each claim to the evidence to a per-claim gate. The gate $\sigma$ is their mean. Text regions are found with the EasyOCR detector. All runs use one RTX 4090 with 24\,GB. %
Three seeds per configuration. Intervals in the text are 95\% two-level bootstrap intervals over seeds and targets. Hyperparameters follow one validation rule fixed before any attack: the strongest setting within 5\% of Cover's validation NDCG@10. For \efr, $\gamma=0.1$ was fixed in advance and every trained $\gamma$ is reported.
\subsection{Evaluation Results}
\label{sec:results}
\begin{table*}[t]
\centering\footnotesize\setlength{\tabcolsep}{2.4pt}
\caption{Defense performance across encoders, warm rankers and datasets.}
\vspace{-4mm}
\label{tab:enc_main}\label{tab:sv}
\resizebox{0.9\linewidth}{!}{%
\begin{tabular}{@{}l*{16}{c}@{}}
\toprule
 & \multicolumn{8}{c}{{CLIP}} & \multicolumn{8}{c}{{SigLIP}}\\
\cmidrule(lr){2-9}\cmidrule(l){10-17}
 & \multicolumn{4}{c}{{MicroLens-100K}} & \multicolumn{4}{c}{{ShortVideo}} & \multicolumn{4}{c}{{MicroLens-100K (ID-hybrid)}} & \multicolumn{4}{c}{{ShortVideo}}\\
\cmidrule(lr){2-5}\cmidrule(lr){6-9}\cmidrule(lr){10-13}\cmidrule(l){14-17}
System & NDCG & Cold & Gain & Cert. & NDCG & Cold & Gain & Cert. & NDCG & Cold & Gain & Cert. & NDCG & Cold & Gain & Cert.\\
\midrule
Free & 0.0395 & 0.1149 & 33.8 & -- & 0.0092 & 0.0294 & 31.1 & -- & 0.0530 & 0.1340 & 20.5 & -- & 0.0081 & 0.0352 & 31.6 & --\\
UAT-MC & --  & -- & -- & -- & 0.0088 & 0.0317 & 20.6 & -- & --  & -- & -- & -- & 0.0070 & 0.0388 & 14.9 & --\\
\grip & 0.0317 & 0.1046 & 17.9 & -- & --  & -- & -- & -- & 0.0524 & 0.1276 & 4.8 & -- & --  & -- & -- & --\\
Body router & 0.0374 & 0.160 & 0.0 & 0.0 & 0.0093 & 0.0252 & 0.0 & 0.0 & 0.0510 & 0.165 & 0.0 & 0.0 & 0.0082 & 0.0346 & 0.0 & 0.0\\
\efr, $\gamma{=}0.02$ & 0.0371 & 0.163 & 6.7 & 8.6 & 0.0094 & 0.0247 & 5.6 & 7.9 & 0.0506 & 0.171 & 5.4 & 7.5 & 0.0082 & 0.0351 & 5.0 & 7.1\\
\efr, $\gamma{=}0.1$ & 0.0376 & 0.151 & 19.8 & 29.5 & 0.0093 & 0.0275 & 21.2 & 34.1 & 0.0513 & 0.157 & 17.3 & 27.5 & 0.0081 & 0.0373 & 19.7 & 29.3\\
\bottomrule
\end{tabular}}
\end{table*}

\noindent\textbf{Robustness and utility.}
{The main experiment asks whether \efr\ removes the attack gain without the quality cost of \cref{sec:revisit}. \Cref{tab:efr} shows that it does. At $\gamma=0.02$, REAP gains only 5.4\ points (81\% less than on Free), and the upper end of its 95\% interval, 5.9, lies far below the lower end of 27.6 for Free. Reach grows from 3.3\% to 5.0\% instead of from 1.6\% to 12.8\%. Against Free, 69\% of the 900 target-cohort pairs gain more than 20 points, while no pair gains more than 12 points against \efr\ at $\gamma=0.02$ (\cref{fig:gain_dist}). Cold-start Recall@100 rises to 0.170, above every tower that reads the packaging (at most 0.1485\ in \cref{tab:main}), and the merged NDCG@10 stays within 4\% of the undefended system (0.0451\ against 0.0469), because warm items keep their ranker. The rank-calibrated merge matters: merging by per-user standardized scores instead fills 46\% of the top 10 with cold items and loses 34\% of NDCG@10. A small packaging budget also helps honest creators, since $\gamma=0.02$ and $0.05$ beat the body-only router on cold recall. Overall, \efr\ uniquely balances utility, robustness, and certified resistance to cold-start attacks, without sacrificing the quality of all its recommendations.}

\begin{figure}[t]
\centering
\includegraphics[width=\linewidth]{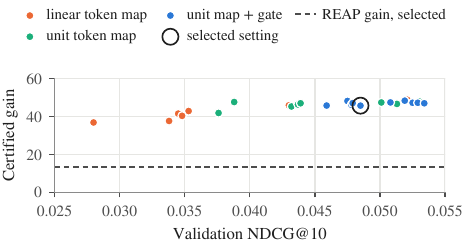}
\caption{Certified gain of the attention bottleneck.}
\label{fig:cert}
\end{figure}

\noindent\textbf{Certification and the choice of $\gamma$.}
{We next test how tight the guarantee is and how $\gamma$ trades exposure for utility. The certificate holds for every packaging edit, not only those REAP tried, and it stays close to the attack. At $\gamma=0.02$ it bounds the gain at 7.5\ points, 1.4\ $\times$ the strongest attack we found, with 31\% of pairs certified below 5 points, and across $\gamma$ the ratio stays between 1.4 and 1.7. The certificate lies above the attack for every pair, and at $\gamma=0.02$ the two correlate at 0.94 (\cref{fig:cert_pairs}). {The bound of the attention bottleneck of Observation 2 is far looser. It certifies 46 to 50 points against an empirical 13 to 15, and none of 25 settings goes below 36.9 (\cref{fig:cert}, details in \cref{sec:exp_cert}).} Raising $\gamma$ increases both gains almost linearly, while cold recall peaks at small $\gamma$ and falls to 0.141\ at $\gamma=0.2$. {Adding the strategic terms of \cref{eq:strategic} barely moves the certified gain, so \efr\ needs no adversarial training.}}

\noindent\textbf{{Generalization across encoders, rankers and datasets.}}
{To test whether the results depend on one setup, we vary the encoder (CLIP), the warm ranker (ID-hybrid) and the dataset (ShortVideo). The pattern holds in all of them (\cref{tab:enc_main}). With CLIP on MicroLens-100K, \efr\ at $\gamma=0.02$ cuts the gain from 33.8\ to 6.7\ and raises cold recall. With the hybrid ranker, strategic training matches its gain at no NDCG cost, but \efr\ keeps higher cold recall and a certificate. On ShortVideo, \efr\ cuts the gain from 31.6\ to 5.0\ at the NDCG@10 of Free, while UAT-MC halves it at a 14\% NDCG@10 cost. The body-only tower does not beat Free on cold items there, so \efr\ buys certified robustness rather than higher accuracy on that dataset.}

\begin{figure}[t]
\centering
\includegraphics[width=\linewidth]{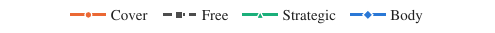}\\[2pt]
\begin{subfigure}[b]{.49\linewidth}\includegraphics[width=\linewidth]{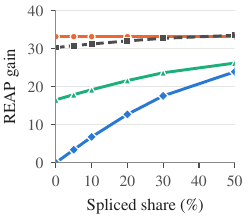}\caption{SigLIP}\end{subfigure}\hfill
\begin{subfigure}[b]{.49\linewidth}\includegraphics[width=\linewidth]{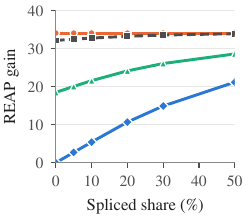}\caption{CLIP}\end{subfigure}
\caption{Adaptive attack by splicing cohort footage.}
\label{fig:adaptive}
\end{figure}

\noindent\textbf{Adaptive attacks on the evidence.}\label{sec:exp_adaptive}
The certificate covers the packaging, not the body, so we test an attacker who knows \efr, keeps the best REAP packaging and replaces a share of the timeline with footage of a popular cohort video (\cref{fig:adaptive}). \efr\ follows the Body curve, which is its own cold path at $\gamma=0$, within $\gamma$: 3.4 points for a 5\% splice and 23.9 for half the timeline. A splice changes what the video contains, so a ranker should respond to it. Telling stolen from new footage is a provenance problem. The bottleneck variant (not plotted) rises from 13.8 to 24.1 points at a 50\% splice, where it meets Body.

\section{Discussion and Limitations}
\label{sec:discussion}
\efr\ rests on the observation that the body of a cold video ranks better than its packaging. This held on both encoders and with content and hybrid warm rankers, but it is a property of the data: where bodies are uninformative, such as static slideshows, a platform would choose a larger $\gamma$, and \cref{prop:efr} states the worst-case exposure this buys. The certificate covers edits of the packaging with the body fixed, and it is per user, so a single cover and title shared by all users stays below it. The same per-user slack is one reason why the bound of the attention bottleneck stays far above the attack, and that bound also relaxes the attention set beyond what fixed keys and unit queries can reach (Remark~\ref{rem:loose}).
Our study has limits. No humans annotated edits, so realizability rests on construction. There is no second-stage ranker. REAP queries exact proxy scores, which favors the attacker. Evidence is frames only. Ablations use a single seed per variant. Cohorts are tag-based and the data come from one platform, with comments standing in as a proxy for the exposure of a video. Extending the evidence to audio and speech is left to future work. %

\section{Conclusion}
\label{sec:conclusion}

This paper revisits packaging attacks in short-video recommendation. We show that creator-realizable edits to covers, titles, and hashtags remain effective after platform processing, while existing adversarial-training defenses struggle to achieve both robustness and recommendation quality.
To address this issue, we propose Evidence-First Ranking (EFR), which explicitly separates creator claims from video evidence, ranks cold items primarily according to the video body, and strictly bounds the influence of packaging. This design preserves recommendation quality while providing a provable guarantee against arbitrary packaging edits.
Our findings suggest that the key challenge in cold-start recommendation is not learning to resist packaging, but avoiding reliance on packaging when allocating exposure.

\section*{Ethics Statement}
All experiments are offline, on public data under its academic license, and never touch a live platform. MicroLens videos are not redistributed. Edited covers stay on the lab server, and the paper shows one cover, taken from a cooking video with no people. We release the edit grammar, the attack and defense code and aggregate numbers. The attack describes edits that creators already make by hand, and its main finding points to a defense that platforms can deploy cheaply. We believe that the defensive value of this work outweighs the small help it gives to an attacker.

{
    \small
    \bibliographystyle{ieeenat_fullname}
    \bibliography{main}
}

\appendix

\section{Implementation Details}
\label{sec:supp_impl}

\noindent\textbf{Edit grammar.}
A program combines any of the following parts. Burned-in text (probability 0.85) uses one of about 280 LLM-proposed captions per cohort (2 to 5 words, \cref{sec:reap}), placed at a uniformly random position, with height 8 to 22\% of the cover and one of four colors with a black stroke. A sticker (probability 0.3) is a square center crop of one of the 50 most-liked covers of the cohort, 20 to 45\% of the cover height. A recolor (probability 0.3) applies a Reinhard-style color transfer toward the palette of one popular cohort cover with strength 0.3 to 1. A crop (probability 0.2) removes up to 15\% on each side. Hashtags (probability 0.85) append three of the 20 TF-IDF keywords of the cohort to the title. The OCR claim of an edited cover is its original OCR text plus the overlay string, verbatim.

\noindent\textbf{REAP.}
The gain among users outside the target cohorts is measured on 1,000 such users.

\noindent\textbf{Recommenders.}
Beyond the setup of \cref{sec:protocol}, the user tower uses feed-forward width 256, dropout 0.2, learned positions and pre-normalization. Warm metrics rank the held-out item against all warm items with the history masked. Cold Recall@100 ranks the held-out cold item against the whole video catalogue of the dataset.

\noindent\textbf{Baselines.}
AMR adds a single-step feature perturbation of $\ell_2$ norm $\epsilon$ on the claim features of all warm items, along the gradient of the recommendation loss, and trains on the clean and perturbed losses. UAT-MC follows Algorithm 1 of the authors: the perturbation of step $k$ builds step $k+1$, the maximization adds $\alpha$ times the cosine between the visual and textual gradients, computed with a double backward pass, and the minimization adds $\lambda$ times the adversarial loss. We searched $\epsilon\in\{0.02,0.05,0.1\}$ for both and $\alpha\in\{1,5\}$ for UAT-MC with $\lambda=1$, and applied the selection rule of \cref{sec:protocol} separately to each tower. On the Cover tower it chose $\epsilon=0.02$ for both methods. On the Free tower, which has more quality headroom, it chose $\epsilon=0.1$, the setting used in the main text, with $\alpha=1$. Cover text masking runs the same CRAFT-based detector as the evidence pipeline on the channel output of every candidate and fills detected regions with a blurred background before encoding. Titles and hashtags are not masked.

\noindent\textbf{Selection of strategic training.}
For strategic training we trained $\lambda_1\in\{1,2,3,5\}$ with $\lambda_2=0.5$. The values $\lambda_1=3$ and $5$ fall below the validation floor, so the rule selects $\lambda_1=2$. For the bottleneck variant we varied the cap $B\in\{1,2,4\}$, the residual weight $\gamma\in\{0.05,0.1,0.2,0.5,1\}$, the token map, the gate and $\lambda_1$. With a linear token map every setting fell outside the utility margin, which led to the unit token map. Hybrid towers use ID dropout 0.1, selected for Cover by the same rule.

\section{Strategic Training Baselines}
\label{sec:supp_grip}\label{sec:grip}
This appendix details the strategic training baseline of Observation 2 (\cref{eq:strategic}) and its variant with an evidence bottleneck. Strategic training changes how a content tower is trained, not what it looks at. It keeps an unconstrained tower and adds two terms to the recommendation loss: one that makes the tower indifferent to the creator's best realizable response, and one that keeps it responsive to real improvement. \Cref{sec:eb} describes the bottleneck variant, which also restricts the architecture and admits a formal bound. In tables, ``\grip'' denotes the first variant and ``Strategic + bottleneck'' denotes the second.

\subsection{Tower and Objective}
\label{sec:tower}
The tower $f$ maps the claims, namely the cover embedding $c_{\mathrm{img}}$, the embedding of the title with its hashtags $c_{\mathrm{txt}}$ and the embedding of the cover's OCR text $c_{\mathrm{ocr}}$, together with the mass-weighted body mean $\bar v=\sum_k m_k v_k$, to a unit-length item vector $z$ as follows,
\begin{equation}
z=\frac{f([c_{\mathrm{img}};c_{\mathrm{txt}};c_{\mathrm{ocr}};\bar v])}{\|f(\cdot)\|},
\label{eq:free}
\end{equation}
with a two-layer MLP $f$ of width 512 and output dimension $d=128$. Without the extra terms this is the \emph{Free} baseline of \cref{sec:protocol}. A SASRec-style user tower~\cite{kang2018sasrec} encodes the last 20 items of a history into a unit vector $h(u)$, and scores are $s(u,i)=\langle h(u),z_i\rangle$. The training loss is
\begin{equation}
\Lo=\Lo_{rec}+\lambda_1\Lo_{inv}+\lambda_2\Lo_{imp},
\label{eq:loss}
\end{equation}
where $\Lo_{rec}$ is the full-softmax next-item cross-entropy over all warm items (\cref{sec:protocol}). In the hybrid form, a learned ID vector is added to $z$ for warm items and dropped with probability $p$ during training, so the content path also learns to stand alone for cold items.

\subsection{An Orbit of Realizable Best Responses}
\label{sec:orbit}
The invariance term needs gaming edits whose effect on the current model is as large as possible. Running REAP inside every training step is too slow, so we precompute an \emph{orbit} and let the model pick from it.

\noindent\textbf{Construction.}
We sample 4,000 warm items with 3 to 200 training interactions. Cold items never enter the orbit. Each item receives two cohorts other than its own tag, and for each pair we draw 12 programs from the REAP grammar of \cref{sec:reap}, render them through one channel draw, and encode the result. This gives 8,000 entries with 12 candidates each, about 96,000 rendered edits per encoder. Every entry is gaming by construction: the body is untouched and the claims point to an unrelated cohort. No judge decides what counts as gaming in this orbit.

\noindent\textbf{Best response.}
Every $R=3$ epochs, the current model encodes all candidates and keeps, for each entry, the one with the highest mean score over 1,000 training users of the target cohort. The loss of the following epochs uses these best responses. This is a best response over a fixed candidate set rather than the full search used for evaluation, and the gap between the two is the price of tractable training. In \cref{sec:exp_ablation}, drawing a random candidate instead leaves more of the attack gain on both architectures, though the difference stays within the seed spread.

\subsection{Two-Sided Invariance}
\label{sec:inv}
Let $a^\star$ be the best response of entry $(i,g)$ and $U_g$ the 1,000 training users of cohort $g$ that also select the best response. The invariance term asks that the mean score of the cohort does not move under the edit:
\begin{equation}
\Lo_{inv}=\E_{(i,g)}\Big|\,\frac{1}{\tau|U_g|}\sum_{u\in U_g}\big(s(u,a^\star(i))-s(u,i)\big)\Big|.
\label{eq:inv}
\end{equation}
Two simpler choices fail, and the failures are instructive.

\noindent\textbf{Why not a listwise KL?}
A natural alternative matches the softmax of cohort scores before and after the edit. This distribution is invariant to adding a constant to every user's score, and a constant rise across the cohort is exactly what lifts an item in everyone's ranking, so the KL term cannot see the dominant effect of gaming. In our ablation it leaves more gain than the two-sided term on both of the architectures we test (\cref{sec:exp_ablation}).

\noindent\textbf{Why not a one-sided penalty?}
Penalizing only a rise, $(\cdot)_+$ instead of $|\cdot|$, looks like the natural objective. The tower can meet it by learning an \emph{edit detector}, which lowers the score whenever the cover or title looks edited. Such a model reports a negative attack gain, which reads as a perfect defense but only means that edits are punished. The two-sided term removes this escape: it asks the tower to ignore the edit, not to punish it. \Cref{sec:exp_ablation} measures the one-sided variant on attacks and on honest edits. On development items, an honest creator who repeats words of their own title on the cover loses 39.6 rank-percentile points among real commenters under the one-sided variant of \grip, against 6.7 under the two-sided term of \cref{eq:inv} on the same development items (\cref{tab:probe}).

\subsection{Preserving Improvement}
\label{sec:imp}
An invariance term alone pushes towards the Body model, which ignores claims entirely. The improvement term keeps two kinds of legitimate effort rewarded:
\begin{equation}
\begin{aligned}
\Lo_{imp}={}&\E\Big[\big(\mu-\langle\bar h_g,z(C,V^{+})-z(C,V)\rangle\big)_+\Big]\\
&+\E\Big[\big(\langle\hat e,z_{\mathrm{rnd}}-z_{\mathrm{rep}}\rangle\big)_+\Big].
\end{aligned}
\label{eq:imp}
\end{equation}
In the first term, $V^{+}$ replaces the second half of the item's timeline with footage of a popular video of cohort $g$, carrying half the temporal mass, and $\bar h_g$ is the mean cohort user vector. A real change of content must move the item towards the cohort by at least the margin $\mu=0.1$. In the second term, $z_{\mathrm{rep}}$ uses the body frame closest to the body mean as the cover and $z_{\mathrm{rnd}}$ a random body frame, and $\hat e$ is the normalized item vector computed from the body alone, the direction of the item's own content. A representative cover must not score below a random one. The margin is what separates strategic training from adversarial training, which flattens all responses inside a budget whether or not those responses are legitimate.

\subsection{Variant with an Evidence Bottleneck}
\label{sec:eb}
Strategic training reduces gaming only for edits like those in the orbit. The bottleneck variant also limits what any claim can do by construction. Claims select and re-weight what the body already contains, and add to it only through a small residual whose norm is bounded.

\noindent\textbf{Capped attention.}
An MLP maps the claims to $L=4$ unit queries $q_\ell$, and a linear map gives unit keys $K_k$ for the evidence tokens. The logits are capped:
\begin{equation}
l_{\ell k}=B\,q_\ell^\top K_k,\quad |l_{\ell k}|\le B,\quad
\alpha_k=\frac1L\sum_{\ell=1}^{L}\frac{m_k e^{l_{\ell k}}}{\sum_j m_j e^{l_{\ell j}}}.
\label{eq:attention}
\end{equation}
The temporal mass $m_k$ enters as a prior, so a token that covers one second of a thirty-second video starts with a thirtieth of the attention. The item vector is
\begin{equation}
\begin{aligned}
&z=\sum_{k}\alpha_k\,\phi(v_k)+\gamma\,\sigma(c,V)\,r(c),\\
&\|\phi(v_k)\|=1,\quad \|r(c)\|\le 1,
\end{aligned}
\label{eq:item}
\end{equation}
where $\phi$ is an MLP followed by normalization. Unit tokens cap $\|z\|$, which stops the norm from encoding popularity, a signal cold items cannot have. The residual $r$ carries claim information without a visual counterpart, such as the language of the title. The support gate $\sigma\in[0,1]$ is a small network on the cosines between each claim and the evidence tokens in the shared encoder space, so a claim that matches nothing in the body is silenced. The vector $z$ is not normalized, so that the bound below is a bound on the score itself. At $B=0$ the attention equals the masses and claims act only through the residual. At $B=0$ and $\gamma=0$ they have no effect and $z$ is the mass-weighted mean of $\phi(v_k)$. As $B$ grows the attention becomes free.

\begin{figure*}[t]
\centering
\includegraphics[width=\linewidth]{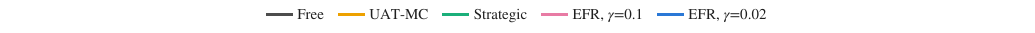}\\[2pt]
\begin{subfigure}[b]{.49\linewidth}\includegraphics[width=\linewidth]{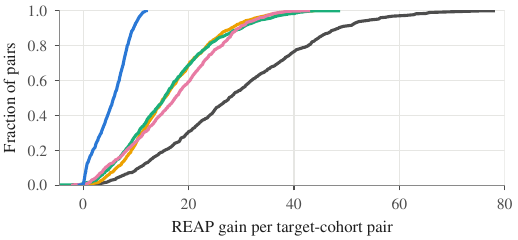}\caption{Distribution over 900 pairs}\end{subfigure}\hfill
\begin{subfigure}[b]{.49\linewidth}\includegraphics[width=\linewidth]{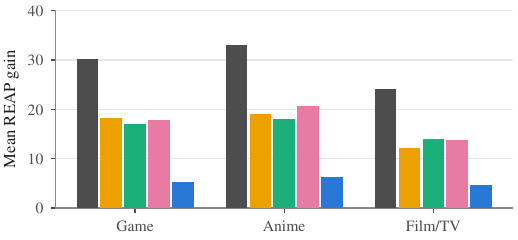}\caption{Mean per target cohort}\end{subfigure}
\caption{REAP gain per target-cohort pair on SigLIP. (a) Cumulative distribution over 900 pairs. (b) Mean gain per target cohort.}
\label{fig:gain_dist}
\end{figure*}

\begin{figure*}[t]
\centering
\begin{subfigure}[b]{.49\linewidth}\includegraphics[width=\linewidth]{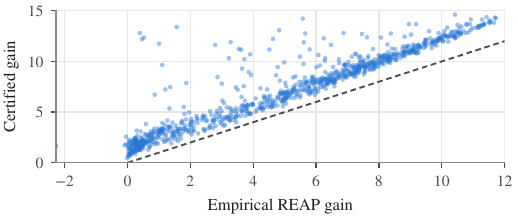}\caption{\efr, $\gamma=0.02$}\end{subfigure}\hfill
\begin{subfigure}[b]{.49\linewidth}\includegraphics[width=\linewidth]{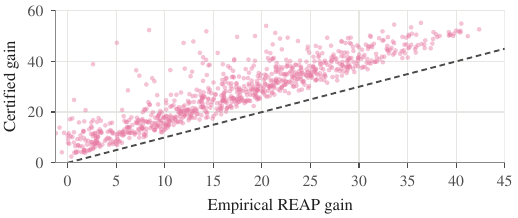}\caption{\efr, $\gamma=0.1$}\end{subfigure}
\caption{Certified against empirical REAP gain per target-cohort pair on SigLIP.}
\label{fig:cert_pairs}
\end{figure*}

\noindent\textbf{A bound over all claims.}
Write $a_k=\langle h(u),\phi(v_k)\rangle$ and sort the tokens with $m_k>0$ so that $a_1\ge\dots\ge a_M$.

\begin{proposition}[Claim bound]
\label{prop:bound}
For fixed evidence $(V,m)$ and every unit $h(u)$, every claim $c$ satisfies
\begin{equation}
\begin{aligned}
s(u,i)&\le\bar s(u)=\max_{0\le p\le M}\frac{N_p}{D_p}+\gamma,\\
N_p&=e^{B}\sum_{k\le p}m_k a_k+e^{-B}\sum_{k>p}m_k a_k,\\
D_p&=e^{B}\sum_{k\le p}m_k+e^{-B}\sum_{k>p}m_k.
\end{aligned}
\label{eq:bound}
\end{equation}
The first term is the exact maximum of $\sum_k\alpha_k a_k$ over the set $\mathcal{S}$ defined in the proof, which contains every attention vector the architecture can produce. It is therefore an upper bound, attained only if the keys allow it.
\end{proposition}
\begin{proof}
Let $\mathcal{S}=\{\alpha\in\Delta:\alpha_k m_j\le e^{2B}m_k\alpha_j\ \forall j,k\}$. Each single-query attention in \cref{eq:attention} lies in $\mathcal{S}$, because the ratio of two of its entries divided by the ratio of their masses is $e^{l_k-l_j}\le e^{2B}$. The set $\mathcal{S}$ is cut out by linear inequalities, so it is convex and contains the mean over queries. Conversely, any $\alpha\in\mathcal{S}$ is proportional to some $w$ with $w_k\in[m_ke^{-B},m_ke^{B}]$, so maximizing $\sum_k\alpha_ka_k$ over $\mathcal{S}$ equals maximizing $f(w)=\sum_kw_ka_k/\sum_kw_k$ over this box. At the optimal value $\lambda$, the optimum also maximizes $\sum_kw_k(a_k-\lambda)$, so $w_k$ is at its upper end when $a_k>\lambda$ and at its lower end when $a_k<\lambda$. This is a threshold on the sorted $a_k$, so the optimum is one of the $M+1$ prefixes in \cref{eq:bound}. Finally, $\langle h(u),\gamma\sigma r\rangle\le\gamma$ because $\sigma\le1$, $\|r\|\le1$ and $\|h(u)\|=1$ hold by construction.
\end{proof}

With cumulative sums the bound costs $O(M\log M)$ per user. It holds for arbitrary claims, including strings that no search would try, as long as the evidence is fixed.

\begin{corollary}[Rank certificate]
\label{cor:rank}
Let $i$ be a cold item, so that $h(u)$ and the scores of all other items do not depend on its claims. For every claim, the rank of $i$ for $u$ is no better than one plus the number of catalogue items whose score exceeds the bound $\bar s(u)$ of \cref{prop:bound}.
\end{corollary}

\noindent\textbf{The price of touching the evidence.}
The bound says nothing about edits that change the body, and a creator can splice footage into the video. The temporal mass prices this.

\begin{proposition}[Splice leverage]
\label{prop:splice}
If a set of tokens $S$ has total mass $f$, every claim gives
\begin{equation}
\sum_{k\in S}\alpha_k\le\frac{e^{2B}f}{e^{2B}f+1-f}.
\label{eq:splice}
\end{equation}
\end{proposition}
\begin{proof}
For one query, the attention on $S$ is largest with $l_k=B$ on $S$ and $l_k=-B$ elsewhere, which gives the right side after dividing numerator and denominator by $e^{-B}$. Since the mean over queries cannot exceed the largest single-query value, the same bound holds for $\alpha$.
\end{proof}

At $B=1$, a splice of 5\% of the duration can win at most 28\% of the attention and a splice of 30\% at most 76\%. At $B=2$, a 5\% splice can already win up to 74\%. The cap therefore sets a real trade-off between how much claims can steer and how cheaply footage can be bought.

\begin{remark}[What the bound does and does not buy]
\label{rem:loose}
\Cref{prop:bound} relaxes the problem in two ways. First, it is computed for each user separately: the worst case for user $u$ may put attention on one token and the worst case for $u'$ on another, while a real item has a single cover and title that every user sees. The certified gain of a cohort is therefore a mean of per-user worst cases, which can be far above the gain of any one claim. Second, $\mathcal{S}$ lets every token reach either end of the cap independently, while the keys are fixed by the evidence and the queries are unit vectors, so many vectors in $\mathcal{S}$ are not attainable. The residual adds $\gamma$ for every user on top. \Cref{sec:exp_cert} shows that at every setting we trained the certificate stays far above the empirical attack.
\end{remark}

\subsection{The bound of the evidence bottleneck}
\label{sec:exp_cert}

For each target, cohort and evaluation user we compute the certified rank of \cref{cor:rank}. \Cref{fig:cert} in the main text summarizes all 25 settings. On CLIP the selected setting certifies 48 to 49 points across seeds, and the certified top-500 reach stays above 99.9\%. Lowering $\gamma$ mainly lowers the certified reach, from above 99\% to 21\%, while the certified mean gain never falls below 36.9. The looseness is structural (\cref{rem:loose}): the bound lets every user face a different worst-case claim, while a creator publishes one, and it relaxes the attention set beyond what fixed keys and unit queries can actually reach.

\subsection{Further baselines and attack diagnostics}
\label{sec:supp_results}
\begin{table}[t]
\centering\footnotesize\setlength{\tabcolsep}{2pt}
\caption{Further baselines and attack diagnostics on SigLIP.}
\vspace{-4mm}
\label{tab:supp}\label{tab:main_full}
\resizebox{\linewidth}{!}{\begin{tabular}{@{}lcccccc@{}}
\toprule
Model & NDCG@10 & Cold R@100 & Pool & REAP gain & Cover & Title\\
\midrule
Cover (undefended) & 0.0400 & 0.1006 & 28.4 & 32.0 & 0.7 & -2.0\\
Free (undefended) & 0.0475 & 0.1133 & 26.2 & 29.2 & -0.2 & 2.9\\
AMR, Cover tower, $\epsilon{=}0.02$ & 0.0390 & 0.1248 & 27.1 & 30.4 & 0.7 & -3.4\\
UAT-MC, Cover tower, $\epsilon{=}0.02$ & 0.0388 & 0.1255 & 26.9 & 30.2 & 0.6 & -3.5\\
\midrule
\grip & 0.0409 & 0.1135 & 13.7 & 16.4 & 0.2 & 2.5\\
\grip, $\lambda_1{=}1$ & 0.0455 & 0.1204 & 14.2 & 17.2 & 0.3 & 2.0\\
\grip, population-wide & 0.0426 & 0.1124 & 14.2 & 16.9 & 0.1 & 2.7\\
\bottomrule
\end{tabular}}
\end{table}
\Cref{tab:supp} complements \cref{tab:main} with three seeds on MicroLens-100K. Pool is the REAP gain of the best sampled program before CMA-ES refinement. Cover and Title are the rank effects, for an item's real commenters, of a representative body frame as the cover and of an accurate title. On the Cover tower, AMR and UAT-MC at the budget the validation rule selects there barely reduce the gain (30.4 and 30.2 against 32.0), and the two variants of strategic training, a smaller $\lambda_1$ and invariance over the whole user population, give gains within one point of the selected setting. Sampling alone obtains most of the final gain, and CMA-ES adds about three points. Faithful presentation changes an item's rank for its real commenters by less than two points for the cover and four for the title, which is why we measure improvement with real content changes.

\section{Visual Analysis}
\label{sec:supp_vis}

\noindent\textbf{Where the gain goes.}
\Cref{fig:gain_dist} shows the per-pair view behind \cref{tab:efr}, averaged over 3 seeds on MicroLens-100K. Against Free, 69\% of the 900 target-cohort pairs gain more than 20 points and the tail reaches 78. UAT-MC, strategic training and \efr\ at $\gamma=0.1$ roughly halve the whole distribution but keep a long tail. \efr\ at $\gamma=0.02$ removes the tail. No pair gains more than 12 points and 43\% stay below 5. The reduction holds for each of the three target cohorts.

\noindent\textbf{Certificates per pair.}
\Cref{fig:cert_pairs} compares the certified and the empirical gain pair by pair on seed 0. Every pair lies above the dashed diagonal, so the strongest edit REAP found never exceeds the certificate. The certificate also tracks the attack closely. At $\gamma=0.02$ the two correlate at 0.94 with a median gap of 1.9 points, and at $\gamma=0.1$ at 0.90 with a median gap of 9.5 points.

\begin{figure}[t]
\centering
\includegraphics[width=\linewidth]{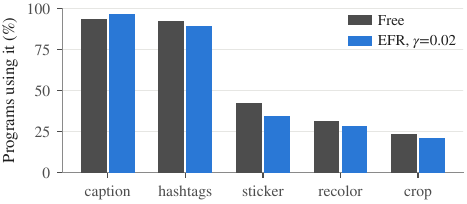}
\caption{Edit operations in the programs selected by REAP.}
\label{fig:edit_mix}
\end{figure}

\noindent\textbf{What the attacker uses.}
\Cref{fig:edit_mix} counts the operations in the programs that REAP selects against Free and against \efr\ at $\gamma=0.02$, with 900 programs each over 3 seeds. Captions and hashtags appear in more than 90\% of them, while stickers, recoloring and crops appear in 21\% to 42\%. Against \efr\ the mix barely changes (captions 97\% against 94\%), so \efr\ does not push the attacker toward other operations. Stickers even become rarer, falling from 42\% to 35\% of the programs, and recoloring falls from 32\% to 28\%. Hashtags drop slightly as well, from 92\% to 90\%. \efr\ bounds the effect of every operation.

\end{document}